# QSAM-Net: Rain streak removal by quaternion neural network with self-attention module

Vladimir Frants[1], Sos Agaian[1], Fellow, IEEE, and Karen Panetta[2], Fellow, IEEE
[1]The Graduate Center, City University of New York, New York, NY 10017 USA
[2] Department of Electrical and Computer Engineering, Tufts University, Medford, MA 02155 USA

Corresponding author: Vladimir Frants (e-mail: vfrants@gradcenter.cuny.edu).

This work was supported in part by the u.s. Department of transportation, federal highway administration (FHWA), under contract 693jj320c000023.

**ABSTRACT** Images captured in real-world applications in remote sensing, image or video retrieval, and outdoor surveillance suffer degraded quality introduced by poor weather conditions. Conditions such as rain and mist, introduce artifacts that make visual analysis challenging and limit the performance of high-level computer vision methods. For time-critical applications where a rapid response is necessary, it becomes crucial to develop algorithms that automatically remove rain, without diminishing the quality of the image contents. This article aims to develop a novel quaternion multi-stage multiscale neural network with a self-attention module called QSAM-Net to remove rain streaks. The novelty of this algorithm is that it requires significantly fewer parameters by a factor of 3.98, over prior methods, while improving visual quality. This is demonstrated by the extensive evaluation and benchmarking on synthetic and real-world rainy images. This feature of QSAM-Net makes the network suitable for implementation on edge devices and applications requiring near real-time performance. The experiments demonstrate that by improving the visual quality of images. In addition, object detection accuracy and training speed are also improved.

## I. INTRODUCTION

Adverse weather conditions (haze, snow, rain, fog) are among the most prevalent reasons for the limited use of automatic video surveillance, crowd counting, accident detection, person re-identification, computational photography, and other computer vision tasks [1-7]. Consequently, a preprocessing step is necessary to limit the impact of weather conditions on the rest of the image analysis system. In real-world applications mentioned above, image de-raining is highly desirable due to the facts that a) rainy weather often causes poor visibility, contrast reduction, and color modification, b) the removal of rain-streaks typically produces over-smoothed images and leads to lost image details, c) the lack of robust prior-models of both rain-streaks and background, and d) the image rain removal is a highly ill-posed problem [8]. Image de-raining aims to create a sharp, clean image from a rainy image. Therefore, it is essential to develop algorithms that automatically remove these artifacts and not degrade the rest of the image's contents. Current deraining efforts can be grouped into model-based and data-driven approaches.

**Model-based** de-raining algorithms make use of the prior information about rain distribution. They exploit images' sparsity and local similarity properties of the rain by using sparse dictionaries and GMM [9-11].

**Data-driven** de-raining methods use deep neural networks trained on end-to-end synthetic and clean/rainy image pairs. Various network architectures were explored, including classical CNNs, recurrent neural networks, and generative adversarial networks [11-15]. Recently, multiscale multi-stage methods such as MPRNet, HINet, and Restormer demonstrated high performance on image processing tasks, including rain streak removal [12, 13, 16].

Despite the significant progress, both model-based and data-driven approaches have their drawbacks. Due to the inadequacy of existing models, some methods tend to remove non-rain-related vertical textures and generate underexposed images [2, 7]. This problem is especially apparent in the case of supervised learning. Modern methods use different network architectures, assumptions, and priors but fail when previously unseen conditions occur [6].

Real-world rainy images are generally diverse and have a complex structure. Raindrops have a variety of sizes, types, densities, and orientations [2, 3, 8]. But the synthetic datasets lack diversity and are limited to the cases of rain of light and mild intensity—Wang et al. attempt to synthesize a real-world dataset using a simple video-based deraining technique [17]. GAN-based methods address the issue by better modeling the dataset distributions and generating images with better illumination, more accurate colors, and better contrast, but tend to produce artifacts when faced with a distribution significantly different from the training [4]. Semi-supervised and unsupervised methods model the rain distribution based on real-world data. This provides tools to generate high-quality synthetic datasets, but many methods fail to show the expected generalization ability, especially in the case of real-world images [5]. Advanced complex deep-learning architectures coupled with infrared imaging, multi-modal input data, and video data improve the situation but are expensive and have limited use in real-world scenarios [8,9]. Bellow, the main limitations of the current state-of-the-art de-raining methods are summarized:

1. *Overfitting, over-smoothing, and unnatural hue change.* Overfitting is caused by the lack of large, diverse training datasets limiting the generalization ability in actual applications [2, 6]. The over-smoothing manifests itself in the textured background regions due to the insufficient rain modeling and the similarity of the rain streaks and background textures [3]. The presence of the mist generally distorts the color information, and many methods cannot restore it [4].

2. *Limited ability to model complex rain patterns, such as multiple overlapping rain streaks and mist.* Streak accumulation over a distance leads to the mist/fog effects[5].

3. *Lack of high-quality realistic datasets.* Data-driven methods are highly dataset-dependent. Therefore, they require extensive training datasets, and the quality of results depends on the dataset's quality [6]. Nevertheless, most of the available training datasets are limited in the size and variation of the available rain distortions.

4. *Lack of focus on consequent processing steps.* Even though CNNs perform well in many real-life applications, they congenitally suffer from shortcomings in color image processing purposes/tasks.

**The attention mechanism:** The attention mechanism has become an essential part of neural networks and their applications, including speech recognition, natural language processing, and computer vision [18]. The goal of the attention module is to a) improve a network expression ability by stressing useful features and suppressing non-important information. Notably, it can effectively suppress unnecessary color features and noise and adaptively enhance the valuable information of the input features, b) advance model training performance, and c) reduce computational complexity [18, 19].

**Motivation for utilizing quaternion-valued neural networks:** Quaternion neural networks (QNN) research has shown distinct improvements over real-valued neural networks and applications in image, speech, and signal processing, image compression, objective image quality assessment, object recognition [20-28] but have not been utilized for image low-level image processing.

Quaternion neural networks are models in which computations in the neurons are based on quaternion numbers. Quaternion number contains one real and three separate imaginary components. This representation is suitable for efficient processing of color (R, G, B) images due to the preservation of color relationships between R, G, and B channels. This is one of the critical limitations of a real-valued CNN (RCNN). Parcollet et al. showed that RCNN fails to capture the color information when trained on a grayscale image, making it impractical in heterogeneous conditions [20].

QNNs reach state-of-the-art performance by reducing the number of training parameters leading to a natural solution to the overfitting phenomenon [27-28]. This makes them promising for real-time applications and implementation on edge devices.

Design challenges and advantages of quaternion neural networks can be found in the survey [26]. In short, the QNNs enable learning of the internal and spatial relations between multidimensional input features. It makes them a good candidate for a basis of low-level image processing methods, including rain removal.

This work aims to exploit the benefits of quaternion image processing and quaternion-valued neural networks in a low-level image preprocessing framework designed to improve the performance of computer vision systems in adverse weather conditions, such as rain and fog.

Our main contributions are:

1) We introduce a novel combined multiscale, multi-stage self-attention QSAM-Net architecture for single image rain streak removal. To the best of our knowledge, it is the first time QNN is used for rain removal.

2) A novel quaternion algebra-based image processing pipeline QSAM-C-Net that enhances image color, visibility, and perceived quality. It handles the range of distortions introduced by poor weather conditions, including rain streaks, mist, fog, and low contrast.

3) Extensive quantative and qualitative experiments show the performance of QSAM-Net compared to real-valued analog and existing state-of-the-art methods in two important ways.

a) The improvement over state-of-the-art up to 1% in PSNR and SSIM, up to 3 points according to SSEQ [29], and up to 2 points according to BRISQUE [30]. It is as shown by extensive evaluation on a diversity of datasets Test100 [31], Test100L [32], Test100H [32], Test1200 [33], and Test2800 [34]. All of this is achieved while having fewer parameters and faster training, as demonstrated in section IV.

b) Experiments show the improvement in accuracy of object detection in poor weather conditions on Rain in Driving (RID) and Rain in Surveillance (RIS) datasets [3] by 1%.

The remainder of the paper is organized as follows. Section II presents an overview of previous work on single-image rain-streak removal and quaternion neural networks for image processing. Section III describes the proposed QSAM-Net architecture. Section IV presents and discusses the experimental results. Section VI concludes the paper.

## II. PREVIOUS WORK

In this section, existing image de-raining methods are discussed as well as recent progress in the area of quaternion neural networks.

### A. SINGLE IMAGE DERAINING

To approach the rain streak removal task, a straightforward model of the rainy image $I$ is represented as a sum of the rain-free scene $J$, and a sparse streak image $S$ [3]:

$$I = J + S$$

Image de-raining is non-trivial for several reasons:
1. It is an ill-posed problem. Since there are many physically accurate interpretations of the same rainy image, it is challenging to develop a method that generates accurate results in various situations [3].
2. Rain comes in many forms, such as drops, streaks, and a combination of rain and mist, requiring separate priors for each case [4].
3. Rain streaks are randomly distributed and have a large appearance variability in size, shape, scale, density, and direction [5].

Many methods have been proposed to address the issues above. One may find a survey on rain removal from video and a single image [8]. The hierarchy of existing methods and their advantages and disadvantages is presented in Fig. 1.

Early investigations into the subject were based on sparse coding, low-rank representations, and the Gaussian mixture models [9-11]. Later, deep-learning-based techniques were introduced [15]. The existing methods can be classified into two main groups: video and single-image rain removal [3]. Video-based methods are highly effective for removing rain-streak patterns in the case of a mostly static background. On the other hand, the problem of single-image rain removal is considered a significantly more complex problem. Below we only consider a single image rain-removal. The recent approaches use advanced network architectures and supervised learning to achieve visually appealing results. Current state-of-the-art methods could be classified into single-stage and multi-stage methods.

**Single-stage approaches:** Fu et al. incorporate long-range contextual information in a dual graph convolutional network to handle the long rain streaks [35]. Wang et al. aim to narrow the domain gap between synthetic rain datasets and real-world rain images by improved rain generation procedure [17]. The method uses GAN for rain generation and implements a Bayesian model parameterized by the physical properties of the rain.

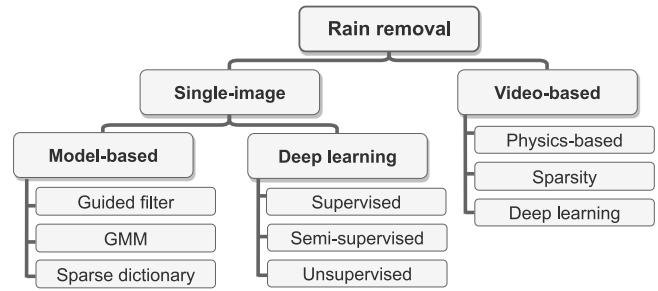

**FIGURE 1.** Hierarchy of existing rain removal approaches.

Specific spatial patterns of rain streaks require the use of multiscale features. Jiang et al. use a coarse-to-fine fusion module to accumulate the streak information across multiple scales with the help of a multiscale pyramid attention mechanism [36]. Restormer is a transformer-based architecture for low-level image processing, including rain streak removal [16]. In many cases, single-stage methods cannot remove the rain streaks completely.

**Multi-Stage approaches** aim to recover a clean image progressively by employing a lightweight subnetwork several times. MPRNet uses a three-stage framework with shared features [12]. Each stage has access to the features generated by the previous step and the input image. HINet investigates the impact of the normalization layer in MPRNet-like architecture on low-level image processing tasks reaching state-of-the-art performance on rain streak removal [13]. Multi-stage approaches deliver excellent performance but are parameter-heavy. It makes them a good starting point for developing a quaternion-algebra-based method.

### B. QUATERNION NEURAL NETWORKS

Quaternion algebra is a valuable tool for multichannel signal processing due to its ability to maintain the relations among dimensions [22]. It prevents information loss and enhances the capabilities of signal processing techniques [23, 22].

Quaternion convolution neural networks (QCNN) use quaternion parameters and redefine the convolution operation using the Hamiltonian product. QCNNs are superior to real-valued CNNs in various image processing and computer vision tasks [28]. The use of quaternion convolution reduces the number of trainable parameters by 75%, simultaneously improving the performance. Sfikas et al. use a QCNN for the segmentation of Byzantine inscriptions. Chen et al. [23] show the superiority of QCNN in image classification and denoising problems. QCNN outperforms a similar real-valued network by 0.89 dB on image denoising and classification accuracy by 4%. Yin et al. [28] derived quaternion batch normalization and introduced a quaternion attention mechanism.

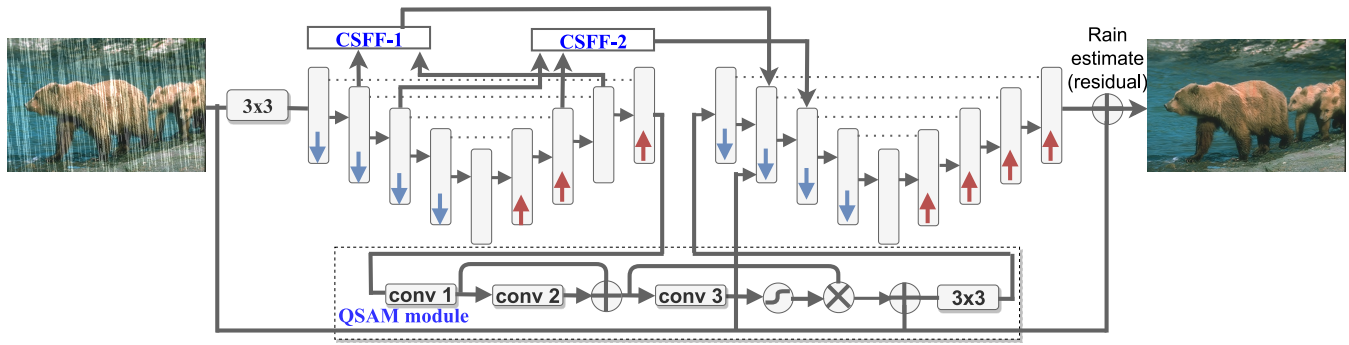

**FIGURE 2.** The proposed quaternion rain-streak removal network QSAM-Net. The network has two stages and aims to estimate the residual rain-streak image $S$ given rainy input $I$. The first stage takes the input image and generates the residual image as well as feature maps with intermediate features after CSFF-1 and CSFF-2, as well as QSAM to generate the input image for the second stage. Both networks have access to the original rainy image.

The resulting QCNN demonstrates superior results on double JPEG compression detection tasks. Parcollet et al. evaluated QCNN on CIFAR-10 and CIFAR-100 classification and KITTI image segmentation dataset [29]. QCNN requires 50% more training time but uses four times fewer parameters and shows slight performance gains. In [27], Parcollet et al. show the ability of quaternion convolutional encoder-decoder (QCAE) to reconstruct a color image from a grayscale input image successfully. At the same time, a similar real-valued network (CAE) fails.

The literature contains a significant amount of evidence to support the claim that using quaternion algebra generally improves performance, especially in the case of color image processing.

## III. PROPOSED QSAM-NET APPROACH

Below we present a quaternion neural architecture QSAM-Net for rain streak removal. Also, we describe a quaternion image processing pipeline QSAM-C-Net for the overall improvement of the visibility of fine details in the image.

### A. QUATERNION IMAGE REPRESENTATION, HAMILTONIAN PRODUCT, AND QUATERNION CONVOLUTION

A quaternion number $\hat{q} \in \mathbb{H}$ extends the concept of complex numbers by introducing one real (a) and three imaginary (b, c, d) components in the form $\hat{q} = a + b\boldsymbol{i} + c\boldsymbol{j} + d\boldsymbol{k}$; where $a, b, c, d \in \mathbb{R}$ and $(\boldsymbol{i}, \boldsymbol{j}, \boldsymbol{k})$ form the quaternion union basis, where $\boldsymbol{i}^2 = \boldsymbol{j}^2 = \boldsymbol{k}^2 = \boldsymbol{ijk} = -\mathbf{1}$.

The color input image of the size W by H pixels is represented as a quaternion matrix $\hat{\mathbf{I}} \in \mathbb{H}^{H \times W}$:

$$\hat{\mathbf{I}} = \mathbf{L} + \mathbf{R}\boldsymbol{i} + \mathbf{G}\boldsymbol{j} + \mathbf{B}\boldsymbol{k}$$

Here $\mathbf{L}, \mathbf{R}, \mathbf{G}, \mathbf{B} \in \mathbb{R}^{H \times W}$ are real-valued matrices representing luminosity, red, green, and blue channels, respectively. Similarly, intermediate feature maps are represented as a group of quaternion-valued matrices. *The main advantage of this image representation* is the preservation of the interrelationship and structural information between the R, G, and B channels [21,22].

An algebra on $\mathbb{H}$ defines operations among quaternion numbers, such as addition, conjunction, and modulus, similarly to the algebra on complex numbers [22]. The Hamiltonian product defines the non-commutative multiplication of two quaternions $\hat{x} = a_1 + b_1\boldsymbol{i} + c_1\boldsymbol{j} + d_1\boldsymbol{k}$ and $\hat{y} = a_2 + b_2\boldsymbol{i} + c_2\boldsymbol{j} + d_2\boldsymbol{k}$ as:

$$\begin{aligned}\hat{x} \otimes \hat{y} = &(a_1 a_2 - b_1 b_2 - c_1 c_2 - d_1 d_2) \\ &+ (a_1 b_2 + b_1 a_2 + c_1 d_2 - d_1 c_2)\boldsymbol{i} \\ &+ (a_1 c_2 - b_1 d_2 + c_1 a_2 - d_1 b_2)\boldsymbol{j} \\ &+ (a_1 d_2 + b_1 c_2 - c_1 b_2 - d_1 a_2)\boldsymbol{k}\end{aligned}$$

In Quaternion Neural Networks (QNN), the Hamilton product ($\otimes$) replaces the real-valued dot product as a transform between two quaternion-valued feature maps. It allows maintenance and exploitation of relations within components of a quaternion feature map. In Quaternion CNN (QCNN), the convolution of a quaternion input $\hat{q} = q_0 + q_1\boldsymbol{i} + q_2\boldsymbol{j} + q_3\boldsymbol{k}$ and kernel $\widehat{W} = W_0 + W_1\boldsymbol{i} + W_2\boldsymbol{j} + W_3\boldsymbol{k}$ is defined as:

$$\widehat{q'} = \widehat{W} \otimes \hat{q}$$

Typically, the quaternion convolution is implemented as a grouped real-valued convolution. A $C$-channel quaternion feature map is represented as $4 \cdot C$-channel real-valued feature map. The first $C$ channels represent real components of quaternion feature maps, and the following three groups of $C$ channels each represent $i$, $j$, and $k$-components. The components of weight $\widehat{W}$ convolved with multiple quaternion inputs.

### B. QUATERNION MULTISCALE MULTI-STAGE STREAK REMOVAL NETWORK

The architecture of the QSAM-Net is presented in fig. 2. We use an encoder-decoder multiscale network with skip connections. Both encoder and decoder are residual blocks with quaternion convolution and quaternion instance normalization layers. We apply two stages consequently. Each stage has access to the original rainy image $I$, and the ground truth image $J$, during the training. The stages are interconnected by Cross-Stage Feature Fusion (CSFF) [12] modified to handle quaternion feature maps and the newly proposed Quaternion Self-Attention Module (QSAM). The network aims to estimate the residual rain-streak image $S$ given rainy input $I$.

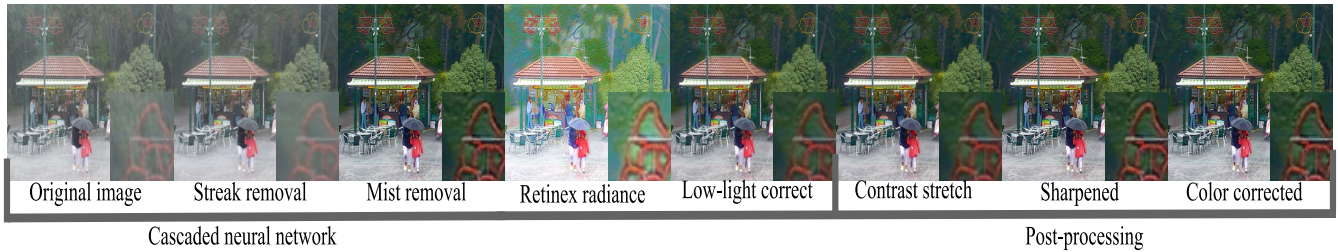

FIGURE 3. The cascaded quaternion network QSAM-C-Net is formed by stacking together a QSAM-Net for streak removal, a modified quaternion GCANet for mist removal, and a Quaternion Retinex decomposition network for radiance estimation and low-light correction with consequent contrast stretching, sharpening and color correction.

For the real-valued case, it is defined as:
$$h(x) = \begin{cases} x & \text{if } x > 0 \\ \lambda x & \text{otherwise} \end{cases}$$

Where $\lambda$ is the gain parameter. We define the quaternion version by splitting along the components of a quaternion as:
$$LeakyReLU(\hat{q}) = h(a) + h(b)\boldsymbol{i} + h(c)\boldsymbol{j} + h(d)\boldsymbol{k}$$

Where $\hat{q} = a + b\boldsymbol{i} + c\boldsymbol{j} + d\boldsymbol{k}$ in an element of input quaternion-valued feature map.

To prevent internal covariate shift, we use Instance Normalization [41], defined as:
$$y_{tijk} = \left(\frac{x_{tijk} - \mu_{ti}}{\sqrt{\sigma_{ti}^2 + \epsilon}}\right)\gamma + \beta$$

Where $\gamma \in \mathbb{R}$ and $\beta \in \mathbb{H}$ are trainable parameters, $(j, k), j = 0..H, k = 0..W$ – pixel coordinates, $i = 0..M$ – feature map index, $t = 0..B$, $B$ – number of images in the minibatch, $M$ – number of feature maps, $H$ and $W$ – are height and width of the feature map, and $\epsilon$ is a small value, added for stability. The mean and variance for the quaternion case are computed as defined in [37].

### C. QUATERNION MULTISCALE MULTI-STAGE STREAK REMOVAL NETWORK

In this subsection, we present a novel quaternion self-attention module (QSAM). It passes the features between the stages of the network. The use of quaternion algebra in the attention module assures better use of multichannel information. The structure and functioning of the QSAM are illustrated in the middle part of fig. 2.

The objective of QSAM is to take the feature map of the previous stage and generate the residual image. Then, the residual is added to the input image to form the restored image used to create the guidance image S. The quaternion self-attention module contains three quaternion convolution operations. First, convolution conv1 enriches the input feature map by using the convolutional kernel of size 3x3. Then the input feature map is transformed to the extent of a single quaternion that is summed up with the input image. Next, this information is used to generate the guidance by the convolution conv3, conv3 after the non-linear operation is used to transform the feature map by the guidance map. In addition to QSAM, we use cross-stage feature fusion modified for the quaternion case [8].

Within the CSFF module, we use 3x3 convolution to transform the features to enrich the multiscale features of the next stage.

### D. IMPLEMENTATION DETAILS

We use $3 \times 3$ convolutional kernels throughout the network. The number of quaternion feature maps in the encoder is set to 16-32-64-128-256. We use PyTorch [38] and Adam [39] with a minibatch of size 2 to train the network. The training patches are of size $256 \times 256$ pixels, with horizontal flip for data augmentation. Starting learning rate is set to $2 \times 10^{-4}$ and decreased down to $1 \times 10^{-7}$ with cosine annealing strategy [35]. We are interested in the impact of the use of QCNN on the streak removal capability, so we optimize the network end-to-end using a very basic mean squared error (MSE) loss function:
$$\mathcal{L} = \frac{1}{HW}\sum_{i=0}^{H-1}\sum_{j=0}^{W-1}[I(i,j) - J(i,j)]^2$$
where $H$ and $W$ – are the height and width of the output image, $I$ is the rainy image, and $J$ is the rain-free ground truth image.

### E. QUATERNION IMAGE VISIBILITY ENHANCEMENT WITH QSAM-C-NET

Most of the papers on single image de-raining ignore the impact of the rain removal on the consequent high-level tasks. Also, rain streaks always coincide with the introduction of the mist, blur, and low-light conditions, leading to poor visibility of fine details in the image. To this end, we describe the QSAM-C-Net visibility restoration network and a postprocessing procedure designed to deal with these issues.

We separately designed and trained the following blocks: (1) QSAM-Net for rain streak removal, (2) a quaternion neural network for mist removal, and (3) a low light image enhancement block. Each block uses quaternion image representation and quaternion algebra for processing. Following, we apply classical image processing methods (1) Image quality enhancement using a histogram adjustment method, (2) sharpening as described in [42], (3) color correction following the procedure from [43]. The pipeline of QSAM-C-Net is illustrated in Fig. 3.

Low-light enhancement using Retinex decomposition is essential for images taken at night and in poor weather

VOLUME XX, 2017

conditions. It also compensates for the general tendency of dehazing and de-raining methods to produce too dark images [44]. The network processes data in three steps: decomposition, adjustment, and reconstruction. The input image is at first decomposed into reflectance and illumination. We only employ the decomposition step to extract the color information in our combined cascaded network. We replace all the convolutional layers with the quaternion convolution layer and replace the ReLU activation layer with its quaternion split version. We trained the network on the same LOL datasets containing 500 low/normal-light image pairs following the procedure from the original paper [44].

To carry out the dehazing operation, we use architecture similar to Gated Context Aggregation Network [45]. We adopted it to the quaternion case. It uses a straightforward and efficient coder-decoder network with half-resolution intermediate features gathered using dilated convolution [46]. We use three encoders, three decoder full-resolution layers, and four half-resolution blocks in our modified network. We keep the network structure. The convolution operation is replaced with the quaternion version. ReLU activation function is replaced with its quaternion split variant. We use the quaternion variant of Instance Normalization, as described in section B. In addition, we keep the gate prediction convolution real-valued. The network is trained on the OTS synthetic dataset [39] for 12 epochs using Adam [34] optimizer with a starting learning rate of 0.01 and decayed by 10 every five epochs. The batch size is set to 12. The hyperparameters are chosen experimentally.

## IV. EXPERIMENTAL RESULTS

This section evaluates QSAM-Net and QSAM-C-Net. First, we compare QSAM-Net to five state-of-the-art methods: SS-IRR [48], MSPFN [18], VRGNet [44], MPRNET [8], HiNet [9] on PSNR and SSIM, and human perceived visual quality metrics (SSEQ [29], BRISQUE [30]) on synthetic and real-world datasets, and high-level computer vision tasks (object detection) on Rain in Driving (RID) and Rain in Surveillance (RIS) datasets [3].

### A. SINGLE IMAGE DERAINING ON SYNTHETIC AND REAL-WORLD DATASETS

We train our model on the synthetic Rain13k dataset [43]. It contains 13712 rainy/rain-free pairs and is composed of previously introduced datasets with various synthetic rain conditions. For testing we use Test100 [31], Test100L [32], Test100H [32], Test1200 [33], and Test2800 [34] datasets that offer a wide variety of synthetic rain for "heavy" and "sparse" cases. MSPFN, MPRNet, and HINET are trained on the Rain13k dataset. SS-IRR and VRG are trained using procedures from the original papers. SSIM and PSNR are computed on the Y channel in the YCbCr color space, which is the standard practice [2].

Metrics BRISQUE and SSEQ model the human visual system's perception but do not consider the color perception and are computed on the greyscale version of the image only. The metrics mainly consider the naturalness and structural features of the images. From the table above, it might be seen that the use of our deraining methods improves the quality measured by SSEQ. Table I and Table II present the evaluation results. We use blind image quality metrics and an internet-images dataset [50] (fig. 4., fig. 7).

TABLE I
COMPARISON WITH PERCEPTUAL METRICS

|  | Internet images | |
|---|---|---|
|  | BRISQEUE | SSEQ |
| Input | 23.914 | 28.723 |
| SS-IRR | 21.150 | 24.222 |
| MSPFN | 26.229 | 28.051 |
| VRGNet | 23.547 | 29.246 |
| MPRNet | 27.923 | 30.931 |
| HiNet | 25.112 | 28.042 |
| QSAM-Net | 24.329 | 21.251 |
| QSAM-C-Net | **20.806** | **18.774** |

### B. REAL-WORLD APPLICATION (OBJECT DETECTION)

Rain in Driving (RID) contains 2495 rainy images taken from real-world driving videos. The dataset includes all three types of rain effects: raindrops on the windshield, mist, and rain streaks. The images are taken in various diverse locations. The dataset is labeled with bounding boxes for 5 types of objects: car, person, bus, bicycle, and motorcycle. Rain in Surveillance (RIS) is similar to RID and contains 2,048 rainy images taken from surveillance videos. The images mostly contain rain streaks and mist. Both datasets are converted to the COCO format, and the resolution of images is changed to 640 on the longer side where necessary [50]. We conduct experiments on a single Nvidia 1080Ti GPU. For evaluation, we use a state-of-the-art network, SCNet trained on COCO dataset [51]. The model uses a multiscale approach to deliver high accuracy in object detection and object segmentation tasks.

We count a detection denoted by a bounding box with a confidence score equal to or higher than 0.3. We do not consider bounding boxes with the predicted classes that are not available in RID/RDS. The detection results are evaluated with a metric proposed for the COCO challenge mAP50 [50]. The examples of detected objects for threshold 0.3 are given in fig. 5. From fig. 4. Table I shows that the quaternion network outperforms HINet and MPRNet using a smaller number of parameters and a very limited amount of training. Generally, our quaternion network has the same or better capability to remove the streaks from real-world images. Fig. 5 shows that our method successfully removes streaks while SS-IRR and VRG fail to remove them completely. Using the proposed cascaded network and postprocessing pipeline improves the visibility significantly.

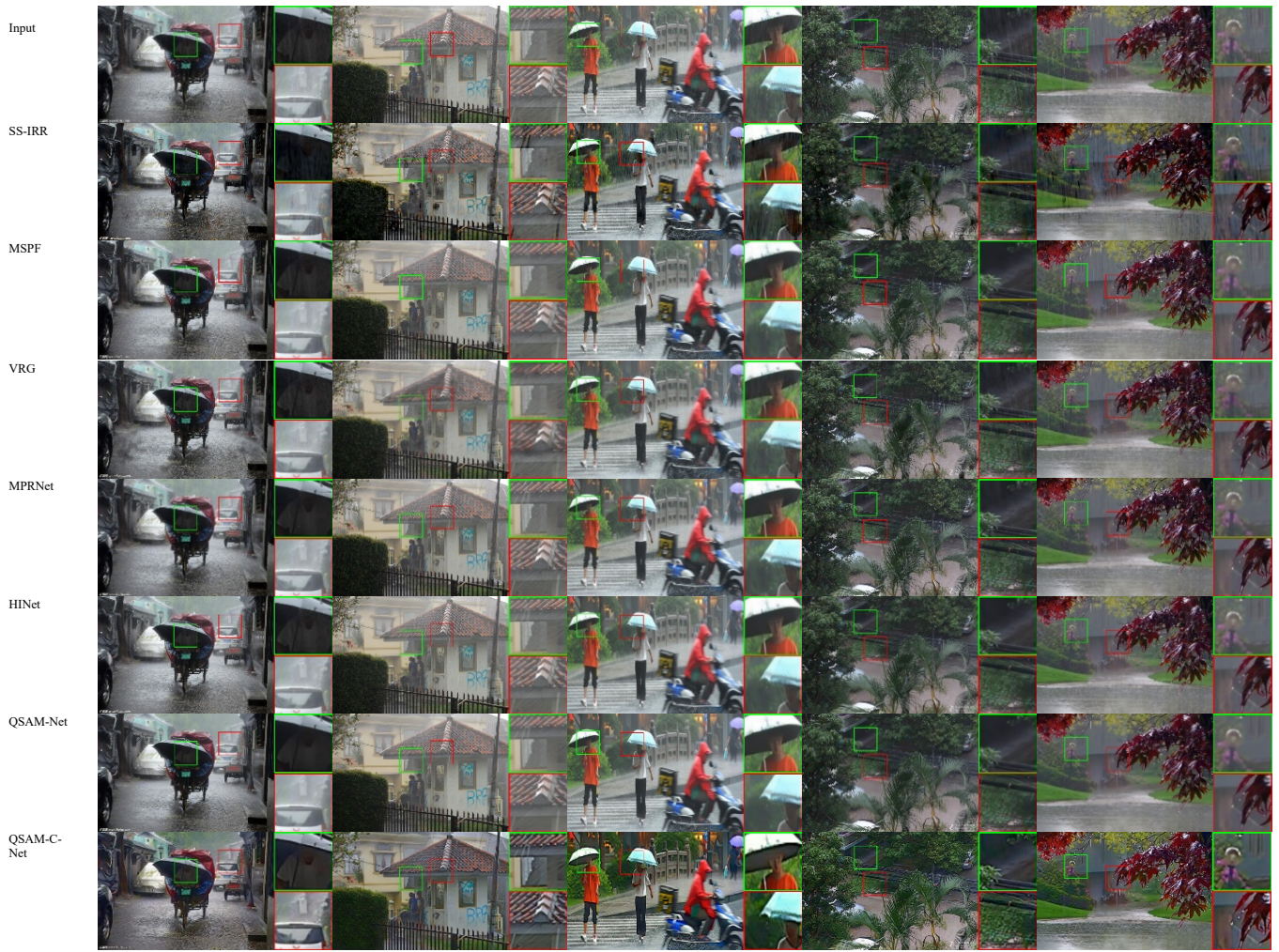

**FIGURE 4.** Rainy images from Internet Images dataset processed with different architectures including SS-IRR, MSPF, VRG, MPRNet, HINet, QSAM-Net, QSAM-C-Net. It can be seen that QSAM-C-Net is qualitatively better at improving visibility than other methods.

TABLE II
COMPARISON RESULTS OF AVERAGE PSNR, AND SSIM ON SEVERAL WIDELY USED RAIN DATASETS, INCLUDING RAIN100H, RAIN100L, TEST100, TEST2800, AND TEST1200. QSAM-NET SHOWS BETTER RESULTS ON BOTH PSNR AND SSIM.

| | Test100 [47] | | Rain100H [48] | | Rain100L [48] | | Test1200 [49] | | Test2800 [50] | | # parameters |
|---|---|---|---|---|---|---|---|---|---|---|---|
| | SSIM | PSNR | SSIM | PSNR | SSIM | PSNR | SSIM | PSNR | SSIM | PSNR | |
| Input | 0.686 | 22.542 | 0.378 | 13.551 | 0.838 | 26.900 | 0.732 | 23.634 | 0.783 | 23.349 | - |
| SS-IRR | 0.787 | 22.261 | 0.490 | 16.458 | 0.842 | 24.892 | 0.824 | 26.097 | 0.867 | 28.072 | 23,200,000 |
| MSPFN | 0.878 | 27.643 | 0.850 | 28.226 | 0.926 | 32.130 | 0.915 | 32.029 | 0.919 | 32.023 | 15,823,424 |
| VRGNet | 0.822 | 23.273 | 0.886 | 30.063 | 0.966 | 34.661 | 0.871 | 28.735 | 0.874 | 28.535 | 170,697 |
| MPRNet | 0.897 | 30.274 | 0.889 | 30.411 | 0.965 | 36.401 | 0.916 | 32.912 | 0.938 | 33.588 | 3,637,303 |
| HiNet | 0.906 | 30.270 | 0.893 | 30.633 | 0.969 | 37.205 | 0.918 | 33.016 | 0.940 | 33.834 | 88,669,702 |
| QSAM-Net | **0.913** | **30.292** | **0.897** | **30.721** | **0.972** | **37.227** | **0.917** | **33.117** | **0.942** | **33.918** | 22,278,819 |

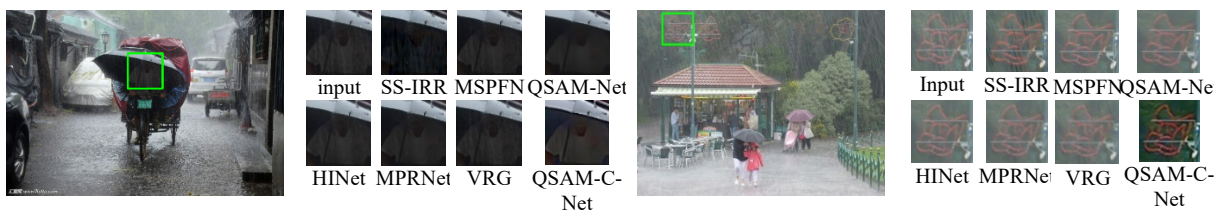

**FIGURE 5.** Dehazing results from various techniques. It can be seen that the presented QSAM-C-Net delivers the best visual quality.

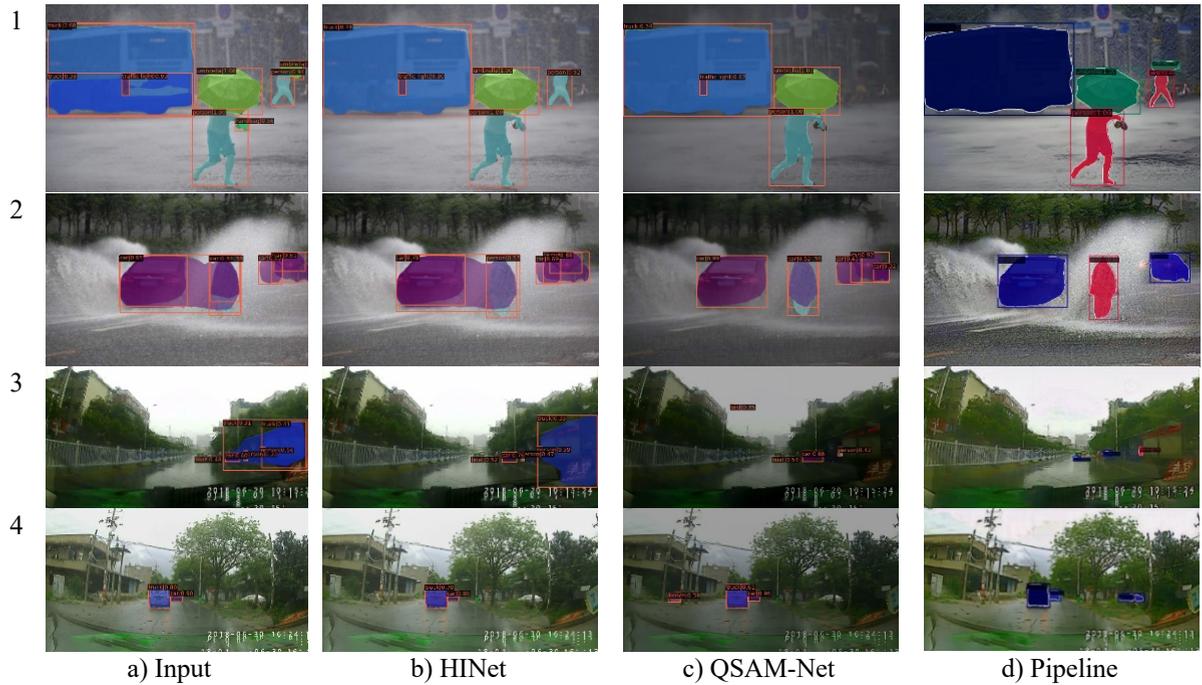

a) Input     b) HINet     c) QSAM-Net     d) Pipeline

**FIGURE 6.** Dehazing results for various techniques. Shows differences in performance of the same detection method depending on the preprocessing procedure used. As could be seen, visibility improvement by the proposed pipeline prevents misdetections and improves the detection of small objects in the background.

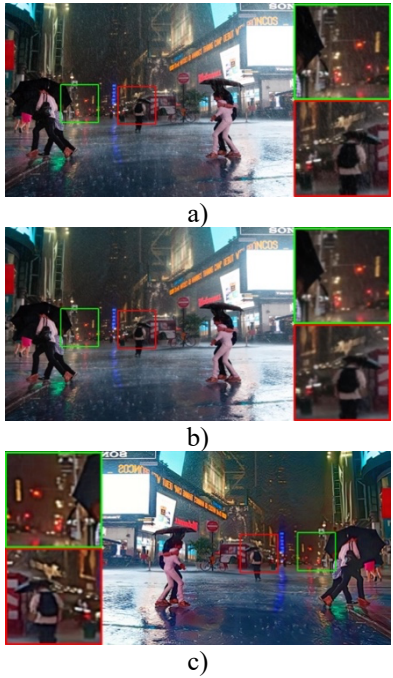

**FIGURE 7.** Image without processing (a), Image processed by QSAM-Net (b) and Image processed by QSAM-C-Net (c)

We compared: the direct use of pre-trained SCNet on the unpreprocessed rainy image and de-raining method followed by SCNet, our network with SCNet, and our pipeline with SCNet. Results for average precision mAP50 are presented in Table IV. Fig. 6 shows examples of detection by the different methods. The results show that preprocessing decreases the accuracy of many methods. This aligns with the previously published results [3].

### C. ABLATION STUDIES

For ablations studies, we use the detection performance as an objective metric of the success of the proposed quaternion pipeline. A comparison of the performance of Quaternion-Valued and Real-Valued networks with the same structure are presented in Table V.

TABLE V
THIS IS A SAMPLE OF A TABLE TITLE

|  | SSIM | PSNR |
|---|---|---|
| Real-valued | 0.933 | 33.075 |
| Quaternion-valued | 0.938 | 33.256 |

From the results above, it is clear that the use of quaternion color representation and quaternion convolution instead of real-valued convolution improves the performance of the dehazing network. The quaternion network trains faster and outperforms the real-valued network at each step during the training, as shown in fig. 8.

TABLE IV
MAP50 FOR RID AND RIS DATASETS

|  | Input | SS-IRR | MSPFN | VRG | MPRNet | HiNet | AM-C-Net |
|---|---|---|---|---|---|---|---|
| RID | 0.300 | 0.274 | 0.300 | 0.290 | 0.298 | 0.302 | 0.314 |
| RIS | 0.298 | 0.276 | 0.298 | 0.288 | 0.298 | 0.314 | 0.323 |

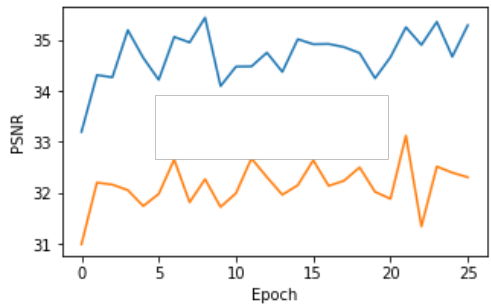

**FIGURE 8.** Test100 testing performance for quaternion (blue) and real-valued (orange) networks trained on Rain13k for 540 iterations, with batch-size 2. Quaternion version of QSAM-Net always has better performance in terms of PSNR.

TABLE VI
THIS IS A SAMPLE OF A TABLE TITLE

| Deraining | Dehazing | Retinex | Sharpening | Color correction | RID mAP | RIS mAR |
|---|---|---|---|---|---|---|
| ✓ | ✓ | ✓ | ✓ | ✓ | 0.305 | 0.325 |
| - | ✓ | ✓ | ✓ | ✓ | 0.302 | 0.315 |
| ✓ | - | ✓ | ✓ | ✓ | 0.298 | 0.308 |
| ✓ | ✓ | - | ✓ | ✓ | 0.304 | 0.322 |
| ✓ | ✓ | ✓ | - | ✓ | 0.305 | 0.324 |
| ✓ | ✓ | ✓ | ✓ | - | 0.304 | 0.324 |

As can be seen, the use of the whole pipeline improves the detection—both de-raining and dehazing impact the detector's performance.

## V. CONCLUSIONS

This paper introduces the QSAM-Net and quaternion image processing pipeline QSAM-C-Net to exploit the benefits of quaternion image processing in the rain removal task. The proposed QSAM module improves the performance by sharing the quaternion features among the stages, building on the Hamiltonian product, and considering the properties of quaternion image representation. QSAM-Net demonstrates up to 1% increase in SSIM and PSNR in various rain conditions on SSIM and PSNR compared to state-of-the-art methods. The proposed quaternion image processing pipeline QSAM-C-Net improves the perceived quality, measured by SSEQ up to 3 points and BRISQUE up to 2 points. Extensive ablation studies show the importance and impact of all the stages in the pipeline. Pariquerly, the proposed neural network (i) shows the benefits of quaternion color representation, quaternion neural networks, and quaternion image processing to ease some limitations of the current methods, (ii) enhances the poor quality of images taken in bad weather conditions while preserving most original details, (iii) improves the performance of object detection., and (iv) approves the importance of rain-streak and mist for automatic image analysis and object detection applications.

We plan to extend the presented method to video-based rain removal using an automatic single image self-learning-based rain streak removal concept in the near future.

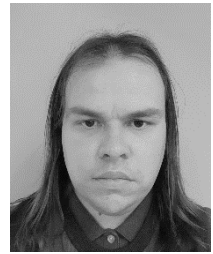

**VLADIMIR FRANTS** received the bachelor's degree (B.E.) degree in electrical engineering from South Russian University of Economics and service, Russia, in 2011, and the M.S. degree in electrical engineering from the Don State Technical University, Russia, in 2013. He is currently pursuing the Ph.D. degree in computer science with The Graduate Center CUNY, USA. His current research interests include artificial intelligence, computer-vision, image processing, and machine learning.

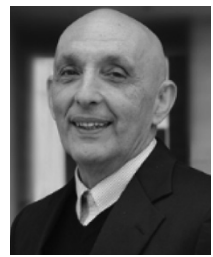

**SOS S. AGAIAN** (Fellow, IEEE) is currently a Distinguished Professor with The City University of New York/CSI. His research interests include computational vision, machine learning, multimedia security, multimedia analytics, biologically inspired signal/image processing, multi-modal biometric, information processing, image quality, and biomedical imaging. He has authored more than 750 technical articles and ten books in these areas. He is also listed as a co-inventor on 54 patents/disclosures. The technologies that he invented have been adopted by multiple institutions, including the U.S. Government, and commercialized by industry. He is a fellow of SPIE, IS&T, AAIA, and AAAS. He received the Maestro Educator of the year, sponsored by the Society of Mexican American Engineers. He received the Distinguished Research Award at The University of Texas at San Antonio. He was a recipient of the Innovator of the Year Award, in 2014, the Tech Flash Titans- Top Researcher Award (San Antonio Business Journal), 2014, the Entrepreneurship Award (UTSA-2013 and 2016), and the Excellence in Teaching Award, in 2015. He is an Editorial Board Member for the Journal of Pattern Recognition and Image Analysis and an Associate Editor for several journals, including the IEEE Transactions On Image Processing, The IEEE Transactions On Systems, Man And Cybernetics, Journal of Electrical and Computer Engineering (Hindawi Publishing Corporation), International Journal of Digital Multimedia Broadcasting (Hindawi Publishing Corporation), and Journal of Electronic Imaging (IS&T and SPIE).

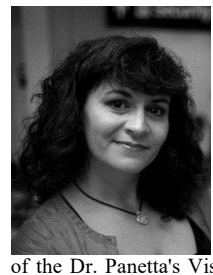

**KAREN PANETTA** (Fellow, IEEE) received the B.S. degree in computer engineering from Boston University, Boston, MA, USA, and the M.S. and Ph.D. degrees in electrical engineering from Northeastern University, Boston. She is currently the Dean of graduate engineering education and a Professor with the Department of Electrical and Computer Engineering. She is also an Adjunct Professor of computer science with Tufts University, Medford, MA, USA, and the Director of the Dr. Panetta's Vision and Sensing System Laboratory. Her research interests include developing efficient algorithms for simulation, modeling, signal, and image processing for biomedical and security applications. She was a recipient of the 2012 IEEE Ethical Practices Award and the Harriet B. Rigas Award for Outstanding Educator. In 2011, she was awarded the Presidential Award for Engineering and Science Education and Mentoring by U.S. President Obama. She is the Vice President of SMC, Membership, and Student Activities. She was the President of the 2019 IEEE-HKN. She is the Editor-in-Chief of the IEEE Women in Engineering Magazine. She was the IEEE-USA Vice-President of communications and public affairs.